\pdfoutput=1

\documentclass[11pt]{article}

\usepackage[]{acl}

\usepackage{times}
\usepackage{latexsym}
\usepackage{linguex}
\usepackage{caption}
\usepackage{graphicx}
\usepackage{subcaption}
\usepackage{array,multirow}
\usepackage{hyperref}
\usepackage{hhline}

\usepackage{stmaryrd}
\usepackage{amssymb}
\usepackage{hhline}
\usepackage{colortbl}
\usepackage{etoolbox}
\usepackage{pgf} 

\definecolor{high}{RGB}{163, 44,12}
\definecolor{low}{rgb}{0.97, 0.97, 1.0}  
\newcommand*{\opacity}{90}
\newcommand*{\minval}{0}
\newcommand*{\maxval}{100}
\usepackage{ifmtarg}
\makeatletter
\newcommand{\isempty}[1]{%
  \@ifmtarg{#1}{YES}{NO}}
\newcommand{\isnotempty}[1]{%
  \@ifnotmtarg{#1}{YES}}
\makeatother
\newcommand{\gradient}[2]{
    \ifdimcomp{#1pt}{>}{\maxval pt}{#1}{
        \ifdimcomp{#1pt}{<}{\minval pt}{#1}{
            \pgfmathparse{int(round(100*(#1/(\maxval-\minval))-(\minval*(100/(\maxval-\minval)))))}
             \xdef\tempa{\pgfmathresult}
            \cellcolor{high!\tempa!low!\opacity}\textcolor{black}{#2}
    }}
}

\usepackage[T1]{fontenc}

\usepackage[utf8]{inputenc}

\usepackage{microtype}

\title{Negation, Coordination, and Quantifiers \\ in Contextualized Language Models}

\author{Aikaterini-Lida Kalouli$^1$\thanks{\hspace{0.1cm} Equal contribution.} ,
  Rita Sevastjanova$^2$\footnotemark[1] ,
  Christin Beck$^2$, \and
  Maribel Romero$^2$ \\
  $^1$~CIS - LMU Munich,  \texttt{kalouli@cis.lmu.de} \\
  $^2$~University of Konstanz,  \texttt{firstname.lastname@uni.kn} \\}

\begin{document}
\maketitle
\begin{abstract}
With the success of contextualized language models, much research explores what these models really learn and in which cases they still fail.~Most of this work focuses on specific NLP tasks and on the learning outcome.~Little research has attempted to decouple the models' weaknesses from specific tasks and focus on the embeddings per se and their mode of learning.~In this paper, we take up this research opportunity: based on theoretical linguistic insights, we explore whether the semantic constraints of function words are learned and how the surrounding context impacts their embeddings.~We create suitable datasets, provide new insights into the inner workings of LMs vis-a-vis function words and implement an assisting visual web interface for qualitative analysis.
\end{abstract}

\section{Introduction}
Recently, tremendous progress has been observed in the development of contextualized language models (LM). After the introduction of contextualized embeddings like ELMo \citep{elmo} and BERT \citep{devlin2019bert},  earlier static word embeddings  \citep{Mikolov2013a, glove} have been sidelined, and new standards have been set for the state-of-the-art. Particularly, LMs have been shown to learn task-agnostic properties of language (e.g., \citealp{belinkov2018,conneau2018-probing,elmo}) and linguistic properties that imitate the classical NLP pipeline \citep{tenney2019-NLP-pipeline}. Despite this success, LMs cannot be taken to understand language the way humans do, as they fail to generalize on unseen data and tasks requiring compositionality \citep{ mccoy-etal-2019-right, richardson2019}.  

Research efforts concentrate on shedding light on what these models learn and consequently how they can be improved. One strand of work relies on probing/diagnostic tasks, where classifiers are trained on LMs to determine whether they can encode specific linguistic properties, e.g.,\ \citet{marvin-linzen-2018-targeted, tenney2019-probing, hewitt-manning2019-probing, talmor2020olmpics}. Another strand of research focuses on creating adversarial test sets with hard linguistic phenomena to observe where the models fail and thus reverse-engineer them, e.g.,\ \citet{mccoy-etal-2019-right, richardson2019, nie-etal-2020-adversarial}.  Both strands of research have approached their goals by employing specific tasks, e.g., Natural Language Inference, Question-Answering, etc., and this has revealed weaknesses of the current models, e.g., that negation is not treated according to its semantic nature. Nevertheless, there is still little work, e.g.,\ \citet{ethayarajh-2019-contextual, sevastjanova-etal-2021-explaining}, decoupling the weaknesses of LMs from specific NLP tasks and focusing on which of their inner workings are responsible for these weaknesses.

In this paper, we attempt to fill this gap by taking a closer look at function words, i.e., words with little lexical meaning. Function words have been traditionally ignored in NLP, being dismissed as stopwords. Since the rise of contextualized embeddings, function words are not dismissed anymore but do not receive any special treatment either; instead, they are contextually learned like any other word. However, the linguistically-motivated datasets  created in this work and tested within the masked-language-modeling (MLM) task show that this context-based learning does not effectively capture the functionality of such words and, thus, that the weaknesses that LMs show in semantic tasks can be attributed to these ill-learned representations. Particularly, our work sheds light to the type and quality of masked-word predictions when \textbf{negation}, \textbf{coordination}, and \textbf{quantifiers} are involved. This work focuses on English masked LMs, but the basic findings should be extendable to other languages since the training process of most masked LMs is relatively similar. 

With this, the contributions of this work are four-fold. First, we show the semantic constraints of three of the most studied classes of function words -- negation, coordination, and quantifiers. 
Second, we provide linguistically-motivated datasets that capture the semantic constraints of these function words. Based on these datasets, we offer new quantitative and qualitative insights into the treatment of function words in masked LMs.~The qualitative insights are facilitated by an openly-available web interface that visualizes LM predictions and allows researchers to easily test their hypotheses.

\section{Related Work}
Most related work, i.e., probing studies and adversarial testing has focused on specific NLP tasks and on the learning outcome of the models. However,  recently, research has also concentrated on exploring the learning procedure per se and understanding the LM inner workings as per the way of their training. With this, the study of contextualization has emerged. The embeddings are learned based on the co-occurrence of words in particular contexts. Thus, the LM embeddings are contextualized, i.e., a word has different representations based on the context it is found in. Particularly, \newcite{ethayarajh-2019-contextual} shows that the embeddings become more contextualized, i.e., more context-specific, in the upper layers of BERT, and that contextualization is not entirely driven by polysemy: (non-polysemous) stopwords such as `and', `of', `the' and `to' also 
become increasingly contextualized in the upper layers, and their representations are among the most context-specific ones. Rather, he finds that contextualization seems to be driven by the variety of contexts a word appears in. These findings are further explored by \citet{sevastjanova-etal-2021-explaining}, who show that contextualization is  neither  driven by polysemy nor by pure context variation, but by the combination of functionality, sense variation, syntactic variation, and semantic context variation: BERT can efficiently model polysemous, homonymous and monosemous words, and also words that appear in semantic contexts of high and low variation and independently of their polysemy. But it cannot model words that have a semi-functional/semi-content nature (e.g., modals, quantifiers, temporal adverbials). In this work, we take up this research direction and further explore what is exactly learned during the learning of function words and how their surrounding context influences their learned embeddings.

\section{Negation,~Coordination,~and~Quantifiers}

\subsection{Theoretical Linguistics}
\label{sec:theory}
Traditionally in theoretical linguistics, there is a distinction between function and content words and several criteria to distinguish the two. The probably most popular criterion is that of semantic content: content words establish a specific semantic content and contribute to the principal meaning of a sentence; function words are rather ``non-conceptual" and mainly fulfill some grammatical function, gluing content words together. Other criteria include membership openness (i.e., whether new members can be added to each of the two classes), the flexibility of syntactic attachment (i.e., whether they can combine with any syntactic phrase), and separability from complements (i.e., whether they can be detached from their lexical head) \citep{corver2013}. The distinction between the two classes is not always clear-cut as there are words that share both functional and lexical properties. Thus, it has often been argued that function and content words should rather form a quasi-continuum
(`squishiness') \cite{ross1972category,emonds1985}. For example,  prepositions are less functional than articles, e.g.,\ some prepositions are associated with a locative or directional meaning, but they are also more functional than nouns or verbs, e.g., because they are inseparable from their content words. 

In this work, we focus on three types of function words, which represent core notions of logic, mathematics, and human reasoning and ones that a state-of-the-art NLP system should efficiently be able to handle: the boolean notions of complement (negation), intersection (conjunction) and union (disjunction), and the notion of existential and universal quantification. Specifically, we study the negation markers \textit{not} and \textit{without}, the coordination markers \textit{or} and \textit{and}, and the existential quantifier \textit{some} and the universal quantifiers \textit{all} and \textit{no}.

\paragraph{Negation Markers} \textit{Not} can be considered syntactic negation while \textit{without}, being a preposition with some lexical content, can be considered lexical negation.
Semantically, at the heart of negation lies the notion of inconsistency \cite{ladusaw1996}: A given sentence and its negation, as in the episodic pair \ref{ex:neg1} and the generic pair  \ref{ex:neg2}, are inconsistent with each other, i.e., they have disjoint truth conditions (where e.g. \textsc{mother} is short for $\{x: x$ is a mother$\}$ and $\overline{A}$ is the complement of $A$):

\ex. \label{ex:neg1} Maria is a mother. \hfill m $\in$ \textsc{mother} \\
Maria is not a mother. \hfill m $\in \overline{\textsc{mother}}$

\ex. \label{ex:neg2} A mom is a mother. \hfill \textsc{mom} $\subseteq$ \textsc{mother}\\
A mom is not a mother. \hfill \textsc{mom} $\subseteq$ $\overline{\textsc{mother}}$

Additionally, negation in natural language interacts with its clausemate elements for meaning composition in interesting ways. To name one case, negation is a well-known ``hole'' for presuppositions, i.e., it can negate the at-issue, propositional content of its clause but not its presupposed content \cite{karttunenpeters1979}. For example, the sentence \textit{Joe isn't sick with covid again} denies that Joe has covid now (at-issue content), but it does not deny that he had covid before (presupposed content).
In this paper, we concentrate on the notion of inconsistency or disjointness and test whether current LMs have grasped this semantic functionality.

\paragraph{Coordination Markers}
Moving on to the coordination markers \textit{and} and \textit{or}, the former is used to form conjunctions and the latter disjunctions. In its run-of-the-mill boolean use, \textit{and} semantically amounts to the intersection operation: Predicating a conjunctive property of an individual, e.g. Joe, amounts to asserting membership to the intersection of the two conjuncts, as in \ref{ex:conj}. 
In contrast, though the purely semantic content of {\it A or B} amounts to set union and thus to inclusive disjunction (meaning `A, B or both'), this meaning is typically strengthened to that of \emph{ex}clusive disjunction (i.e., `A or B and not both') \cite{horn1972,sauerland2004}. This is linked to Hurford's observation that a disjunction of shape {\it A or B} is odd if there is an entailment relation between A and B, as defined in \ref{def:hurford} and illustrated in \ref{ex:disj} \cite{hurford1974, singh2008, ippolito2020}:

\ex. 
    \a. Joe is a dolphin and a mammal. \label{ex:conj} \\
         j $\in$ \textsc{dolphin} $\cap$ \textsc{mammal}
    \b. \# Joe is a dolphin or a mammal. \label{ex:disj} \\
          j $\in$ \textsc{dolphin} $\cup$ \textsc{mammal} $\wedge$ j $\notin$ \textsc{dolphin} $\cap$ \textsc{mammal}

\ex. \label{def:hurford} Hurford's Constraint:\\
    A disjunction of the form {\it A or B} is infelicitous (i.e., \#) if A entails B or vice-versa. 

This means that {\it and} and {\it or} have different semantic signatures when it comes to the relation between their coordinated terms: {\it A and B} allows for A to entail B, whereas {\it A or B} prohibits it. We will test whether LMs are able to detect and learn these different meaning signatures.

There are other interesting semantic properties of coordination markers in natural language that could be explored. To name just one, {\it and} has, in addition to its boolean meaning, a non-boolean meaning that roughly amounts to the creation of a new complex plural individual \cite{link2002}. For example, \ref{ex:and-readings} allows not only for the boolean, distributive reading \ref{ex:and-distr} and also for the non-boolean, collective reading \ref{ex:and-coll}. 

\ex. Jane and Paul built a castle. \label{ex:and-readings}
    \a. `Jane built a castle and Paul built a castle.'\label{ex:and-distr}
    \b. `Jane and Paul together built a castle.' \label{ex:and-coll}

\paragraph{Quantifiers}
Semantically, quantifiers like {\it all/every}, {\it some} and {\it no} denote relations between the set P coming from the noun phrase headed by the quantifier and the set Q coming from the rest of the sentence \cite{barwisecooper1981}:

\ex. 
     \a. $\llbracket$All Ps are Q$\rrbracket$ = \texttt{1} \hspace{4mm}iff P $\subseteq$ Q
     \b. $\llbracket$Some Ps are Q$\rrbracket$ = \texttt{1} iff P $\cap$ Q $\neq \emptyset$ \label{def:some-weak}
     \c. $\llbracket$No Ps are Q$]\rrbracket$ = \texttt{1} \hspace{3mm}iff P $\cap$ Q = $\emptyset$

Two points follow from these lexical entries. First, {\it all} and {\it no} are polar opposites (i.e., they are contrary --though not complementary-- items \cite{ladusaw1996,cruse2011}). This means that the statements {\it All Ps are Q} and {\it No Ps are Q} are inconsistent with each other. Second, the statement {\it All Ps are Q} technically entails the statement {\it Some Ps are Q} (under the assumption that set P is non-empty). However, the two quantifiers form part of the lexical scale $<${\it some}, {\it several}, ... , {\it many}, {\it most}, {\it all}$>$, where the leftmost item is the weakest, and the rightmost item is the strongest \cite{horn1972}. As it typically happens with scales in natural language, pragmatic implicatures are routinely run on the weaker terms, leading to readings not entailed anymore by the stronger terms. In the case of {\it some}, its original weaker reading `some and possibly all' in \ref{def:some-weak} is upgraded into the stronger pragmatic reading `some but not all' in \ref{def:some-strong}. Under this second reading,  the statement {\it All Ps are Q} does no longer entail the statement {\it Some Ps are Q}; rather, the two are incompatible with each other. 

\ex. $\llbracket$Some$_{str}$ Ps are Q$\rrbracket$ = \texttt{1} iff \\ 
P$\cap$Q $\neq \emptyset$ and P$\nsubseteq$Q \label{def:some-strong}

For an LM understanding the meaning of the quantifiers {\it all}, {\it some}, and {\it no} and for a fixed value of P, the Q-embeddings (i.e., embeddings of the set Q) of the noun phrase {\it All Ps} should be radically different from the Q-embeddings of {\it No Ps}, since they are polar opposites (i.e., contraries). Similarly, the Q-embeddings of the noun phrase {\it All Ps} should differ substantially from the Q-embeddings of the pragmatically strengthened {\it Some Ps}, since they are incompatibles. In other words, in both cases, the Q-embedding of the quantificational noun phrases under comparison should be disjoint.  We will test whether LMs show traces of semantic understanding in being able to create such disjoint predictions.

\subsection{Computational Linguistics/NLP}
Within computational linguistics, function words have mostly been treated as \textit{stopwords}. The term was coined by \citet{luhn1960} to mean very common words that do not add much to the meaning of a text but ensure the structure of a sentence is sound. Historically, one of the main reasons for removing stopwords was to decrease the computational time for text mining \citep{huston2010} and help search engines to give better results. Nowadays, this reason for removing stopwords is not valid anymore since we have computationally more powerful hardware; nevertheless, various NLP tasks such as topic modeling and information retrieval, continue to use the practice as it is often argued to improve performance (e.g., \citet{fan2019,serhad2020}). 

With the rise of distributional semantics and word embeddings, more research focused on the nature of these words and the special treatment required (e.g., \cite{bernardi-etal-2013-relatedness, hermann-etal-2013-bad,linzen-etal-2016-quantificational, tang2016}). 
More recently, the special nature of function words has also interested the LM community. In one of the probing tasks proposed by  \citet{kassner-schutze-2020-negated} it is shown that LMs cannot distinguish between negated
(``Birds cannot [MASK]'') and non-negated
(``Birds can [MASK]'') questions. The researchers insert the negation \textit{not} into the LAMA dataset \citep{petroni-etal-2019-language} and create positive and negative cloze questions. By comparing the predictions of LMs on these pairs, they find that the positive and negative predicted fillers have high overlap and correlation, i.e., models are equally likely to generate true (e.g. \textit{birds can fly}) and incorrect statements (e.g., \textit{birds cannot fly}). High correlation means that the models do not understand negation; correct answers for positive and negative questions are expected to be disjoint. Interestingly, the researchers show that BERT can learn both positive and negative facts correctly if they occur in training but still fails to generalize to unseen (positive and negative) sentences.    

\section{Experiments}
In this work we focus on the three functional categories discussed and set out to complement the existing research on the treatment of function words within contextualized LMs. 
First, we create datasets that can be used to evaluate how LMs have learned to capture the semantic constraints of these functional categories. Second, we generate quantitative insights into the contextualization of function words, which we qualitatively investigate through a user-friendly web interface, which also allows researchers to evaluate their own datasets.

\subsection{Data}
\label{s:data}
Motivated by the experiment performed by \citet{kassner-schutze-2020-negated} on the negated LAMA dataset and the theoretical linguistic research about the type of data needed for this kind of exploration, we create two datasets of different types.\footnote{Can be found under: \url{https://function-words.lingvis.io/}} For the creation of the datasets, we combine existing resources: the family and location relations of the analogy dataset used by \citet{Mikolov2013a}, the English occupations dataset of the European Skills and Competences, Qualifications and Occupations (ESCO) initiative\footnote{\url{ec.europa.eu/esco/portal/download}} and 5 common-sense relations found in ConceptNet \cite{speer2016} (\textit{has-a, used-for, is-a, has-property, capable-of}). Based on these resources, we create templates which are then used to produce the sentences of our datasets. Examples for both datasets can be found in the Appendix. More details are given in the following.     

\paragraph{Inconsistent Dataset} The Inconsistent Dataset aims at revealing whether the predictions in sentences with inconsistent meaning are indeed disjoint. Disjointness is factored in through inconsistent function words, as described in Section \ref{sec:theory}. Parts of the inconsistent sentences are masked and the goal is to detect overlapping predictions within each pair. An overlap means that the LM has not learned the functionality of the inconsistent functional words. The Inconsistent Dataset consists of 1272 pairs and can be split into three subsets: the negation subset (534), the coordination subset (500), and the quantifiers subset (238). 

The negation subset contains pairs with positive and negative sentences, whereby the negation can be syntactic or lexical. Concerning syntactic negation, the subset contains examples with the copula verb \textit{be} and its negation, and transitive verbs and their negation with \textit{not} and \textit{no}. 
As far as lexical negation is concerned, inconsistent sentences are created with the copula verb \textit{be} and transitive verbs and based on the preposition \textit{with-without}. 
The coordination subset contains a single type of inconsistency by using the markers \textit{or} and \textit{and} and masking the second part of the coordination. 
The quantifiers subset has two types of opposing contexts: the first one is formulated through \textit{all-no} and the second one through \textit{all-some}. Both transitive and intransitive sentences are considered.  

\paragraph{Semantic Dataset}
Although the Disjoint Dataset goes beyond the negated LAMA in that it contains an additional type of negation (lexical negation) and two further types of inconsistency (coordination and quantifiers), it can indicate whether the LMs learn the semantic constraints of these functional categories only \textit{indirectly}, i.e., by capturing the predictions' overlap. Thus, a different ``semantic'' dataset is required, on which we can directly judge the correctness of the predictions. This can be made clearer with an example such as \textit{A mother is not a [MASK]}: based on the semantic constraints of negation discussed in Section \ref{sec:theory} and the generic reading `For any $x$, if $x$ is a mom, then $x$ is not a MASK', predicted masks should not be one of the words \textit{mom, grandmother, grandma, granddaughter, bride, wife, woman, niece, stepmother, daughter, aunt, etc.} Another example is the sentence \textit{John was born in Berlin or in [MASK]}, where the predicted masks should not contain the word \textit{Germany}; again, see constraints in Section \ref{sec:theory}. We call these words \textit{forbidden} although they could be valid sentences in some (e.g., figurative) contexts and could theoretically have been seen during LM training. However, since these sentences are logically incorrect and do not represent the prototypical concepts of things, they should not be among the most likely predictions of the LM -- since LMs learn based on the occurrences of things, literal, prototypical meanings should have been encountered more often. Thus,  if the most likely predictions contain the forbidden words, it means that the model has neither learned the functionality of the corresponding operator nor any proper world knowledge, e.g., that a \textit{mother} is the same as a \textit{mom} or that \textit{Berlin} is part of \textit{Germany}.  

The creation of such a dataset is particularly challenging because the examples need to be chosen in a way that the necessary semantics is captured within the words themselves and not based on the co-occurrence of the words. Only in this way can we reliably evaluate whether LMs have learned something about the actual semantic constraints of the functional markers or they simply reproduce common concordances of the training data. This means that an example such as \textit{John lives in Berlin or in [MASK]} is not suitable because any predicted word could be right. Thus, to produce such examples, we systematically create examples related to concepts for which we can selectively define invalid predictions according to the particular concept's characteristics. 
We choose family relations and occupations for syntactic negation, animals and their main body parts and activities for lexical negation, and capital countries and animals and their hypernyms for coordination. For the selected quantifiers, no ``semantic'' examples with forbidden answers could be created because of the very nature of quantifiers, i.e., even the universal \textit{all} is not strong enough to create logical invalid examples in the real world. 
The Semantic Dataset contains a total of 2780 sentences with 187 examples of syntactic and 123 examples of lexical negation and 2470 examples of coordination. 
The size of the dataset is comparable to existing datasets used in other related experiments (e.g., the negated LAMA dataset by \citet{kassner-schutze-2020-negated}).

\subsection{Models}
For our study, we use the \textit{huggingface}~\cite{wolfetal2020-transformers} implementation of three pretrained LMs, BERT \citep{devlin2019bert}\footnote{ \url{https://huggingface.co/bert-base-uncased}}, RoBERTa \citep{liu2019}\footnote{\url{https://huggingface.co/roberta-base}}, and ALBERT \citep{albert}\footnote{\url{https://huggingface.co/albert-base-v2}}, which have been shown to achieve state-of-the-art results on the GLUE, RACE, and SQUAD benchmarks. Our two datasets are input to each of the three models to extract the masked predictions, layerwise attentions, and the word embeddings of all words of each sentence.  
The embeddings are taken from layer 11, as the higher layers of models like BERT have been shown to mostly capture semantic properties, while the last layer has been found to be very close to the actual classification task and thus to be less suitable \citep{jawahar2019-bert-structure, linetal-2019-sesame}. Word piece embeddings are merged to their corresponding word embeddings through averaging. Based on the predictions, the word embeddings, and attentions, we calculate the following: 
\begin{itemize}
    \item the cosine similarity between the embedding vector of the predicted word to each other word of the sentence 
    \item the layerwise average attention of the predicted word to each other word of the sentence
    \item for the Inconsistent Dataset: the overlap of the predictions between the two inconsistent versions of the sentences; overlap in the first returned prediction  (\texttt{overlap@1}), in the first 5 (\texttt{overlap@5}), and first 10 (\texttt{overlap@10}) returned predictions 
    \item for the Semantic Dataset: the percentage of examples containing at least one forbidden word within the 1st prediction (\texttt{forbidden@1}), the first 5 (\texttt{forbidden@5}), and the first 10 predictions (\texttt{forbidden@10}). 
\end{itemize}

\subsection{Web Interface}
In addition to the quantitative results, i.e., prediction overlaps and forbidden predictions, we implement a web interface\footnote{\url{https://function-words.lingvis.io/}} that visually shows the predictions, the layerwise average attentions, and the word similarities. The interface provides qualitative insights into similarity patterns that are common for different prediction outcomes. 

In the interface, the user can select one of the three models (i.e., BERT, ALBERT, or RoBERTa) to explore its predictions. Sentences belonging to one dataset are grouped together, whereby disjoint sentence pairs from the Inconsistent Dataset are placed underneath each other for better comparability. The visualization of predictions consists of multiple elements, e.g., see Figure \ref{fig:syn_neg}. Each prediction of a masked word is displayed as a row in the visualization (i.e., 10 rows for 10 predictions). On the left, we display the prediction's probability through a horizontal bar representation. Next to the probability, we display the sentence tokens visualized as colored rectangles. The color of the rectangle represents its cosine similarity or its layerwise average attention to the predicted word of the particular sentence. The darker the color, the higher the similarity/attention. Note that ALBERT and RoBERTa generally show higher cosine similarities between the predicted and the surrounding context words than BERT, i.e., they are displayed with darker colors on the visualized figures. 
On the right, we display the predicted word. To support the analysis of prediction overlaps as well as prediction of forbidden words, we color the words that overlap or are forbidden, respectively, in red color.

\section{Results and Discussion}
The results of our experiments can be found in Tables \ref{tab:resultsOverlap} and \ref{tab:resultsForbidden}. Table \ref{tab:resultsOverlap} shows the percentages of overlap of the inconsistent predictions within the first @1, @5, and @10 predictions. 
Table \ref{tab:resultsForbidden} shows the percentages of at least one forbidden word being included within the first @1, @5, and @10 predictions of the models.

\begin{table}[!ht]
\begin{small}
\begin{center}
\begin{tabular}{c c c c c}
  \textbf{model} & \textbf{Incons. Dataset} & \textbf{@1} & \textbf{@5} & \textbf{@10} \\ 
  \hline
  \parbox[t]{2mm}{\multirow{4}{*}{\rotatebox[origin=c]{90}{\small{BERT}}}} & 
  {coord}  & {\gradient{50}{50}} & {\gradient{53}{53}} & {\gradient{57}{57}} \\
  & {neg} & {\gradient{46}{46}} & {\gradient{46}{46}} & {\gradient{47}{47}} \\ 
  & {quant} & {\gradient{27}{27}} & {\gradient{38}{38}} & {\gradient{40}{40}} \\  
  \hhline{*{1}{|~}*{4}{|-}}
  & {\textbf{all}} & {\gradient{41}{41}} & {\gradient{45}{45}}  & {\gradient{48}{48}}   \\
  \hline
  \parbox[t]{2mm}{\multirow{4}{*}{\rotatebox[origin=c]{90}{\small{ALBERT}}}} & 
  {coord} & {\gradient{61}{61}} & {\gradient{80}{80}} & {\gradient{81}{81}} \\ 
  & {neg} & {\gradient{42}{42}} & {\gradient{42}{42}} & {\gradient{42}{42}} \\
  & {quant} & {\gradient{27}{27}} & {\gradient{28}{28}} & {\gradient{31}{31}} \\ 
  \hhline{*{1}{|~}*{4}{|-}}
  & {\textbf{all}} & {\gradient{43}{43}} & {\gradient{50}{50}}  & {\gradient{51}{51}}   \\
  \hline
   \parbox[t]{2mm}{\multirow{4}{*}{\rotatebox[origin=c]{90}{\small{RoBERTa}}}} & 
   {coord} & {\gradient{40}{40}} & {\gradient{58}{58}} & {\gradient{63}{63}} \\ 
  & 
  {neg} & {\gradient{38}{38}} & {\gradient{40}{40}} & {\gradient{41}{41}} \\  
  & {quant} & {\gradient{25}{25}} & {\gradient{29}{29}} & {\gradient{31}{31}} \\  
  \hhline{*{1}{|~}*{4}{|-}}
  & {\textbf{all}} & {\gradient{34}{34}}  & {\gradient{42}{42}}   & {\gradient{45}{45}}   \\
  \hline
\end{tabular}
\caption{Inconsistent Dataset: percentage of inconsistent pairs with overlapping predictions within the first \textit{x} predictions. }
\label{tab:resultsOverlap} 
\end{center}
\vspace{-1em}
\end{small}
\end{table}

\begin{table}[!ht]
\begin{small}
\begin{center}
\begin{tabular}{c c c c c}
  \textbf{model} & \textbf{Semantic Dataset} & \textbf{@1} & \textbf{@5} & \textbf{@10} \\ 
  \hline
  \parbox[t]{2mm}{\multirow{4}{*}{\rotatebox[origin=c]{90}{\small{BERT}}}} & {synNeg} & {\gradient{41}{41}} & {\gradient{56}{56}} & {\gradient{60}{60}} \\
  & {lexNeg} & {\gradient{51}{51}} & {\gradient{72}{72}} & {\gradient{73}{73}} \\
& {coord} & {\gradient{32}{32}} & {\gradient{73}{73}} & {\gradient{87}{87}} \\ 
  \hhline{*{1}{|~}*{4}{|-}}
  & {\textbf{all}} & {\gradient{41}{41}} & {\gradient{67}{67}}  & {\gradient{73}{73}} \\
  \hline
  \parbox[t]{2mm}{\multirow{4}{*}{\rotatebox[origin=c]{90}{\small{ALBERT}}}} & {synNeg} & {\gradient{14}{14}} & {\gradient{51}{51}} & {\gradient{60}{60}} \\ 
  & {lexNeg} & {\gradient{38}{38}} & {\gradient{63}{63}} & {\gradient{66}{66}} \\
    & {coord} & {\gradient{6}{6}} & {\gradient{31}{31}} & {\gradient{47}{47}} \\ 
  \hhline{*{1}{|~}*{4}{|-}}
  & {\textbf{all}} & {\gradient{19}{19}}  & {\gradient{53}{53}}  & {\gradient{58}{58}}  \\
  \hline
  \parbox[t]{2mm}{\multirow{4}{*}{\rotatebox[origin=c]{90}{\small{RoBERTa}}}} & {synNeg} & {\gradient{48}{48}} & {\gradient{64}{64}} & {\gradient{71}{71}} \\ 
  & {lexNeg} & {\gradient{21}{21}} & {\gradient{31}{31}} & {\gradient{42}{42}} \\ 
    & {coord} & {\gradient{27}{27}} & {\gradient{68}{68}} & {\gradient{82}{82}} \\ 
  \hhline{*{1}{|~}*{4}{|-}}
  & {\textbf{all}} & {\gradient{32}{32}} & {\gradient{54}{54}}  & {\gradient{65}{65}}   \\
  \hline
\end{tabular}
\caption{Semantic Dataset: percentage of sentences in which there is at least one forbidden prediction within the first \textit{x} predictions.}
\label{tab:resultsForbidden}
\end{center}
\vspace{-2em}
\end{small}
\end{table}

\subsection{Findings}
In the following, we describe findings made using quantitative and qualitative evaluation methods.

\paragraph{Same predictions for inconsistent pairs:} We can reproduce and extend the findings by \citet{kassner-schutze-2020-negated}. The first prediction (\texttt{overlap@1}) of disjoint pairs overlaps in 41\% of the cases for BERT, 43\% for ALBERT, and 34\% for RoBERTa. The first 5 predictions (\texttt{overlap@5}) overlap in 45\% for BERT, 50\% for ALBERT, and 42\% for RoBERTa, while the first 10 predictions (\texttt{overlap@10}) overlap in 48\% for BERT, 51\% for ALBERT, and 45\% for RoBERTa. Specifically, \texttt{overlap@10} is worse in the coordination subset in all 3 models and best in the quantifiers subset. This overlap shows that the models neither learn nor consider the functional nature of these markers.

\begin{figure}[!ht]
\centering
\begin{subfigure}{.5\linewidth}
  \centering
  \includegraphics[scale=0.37]{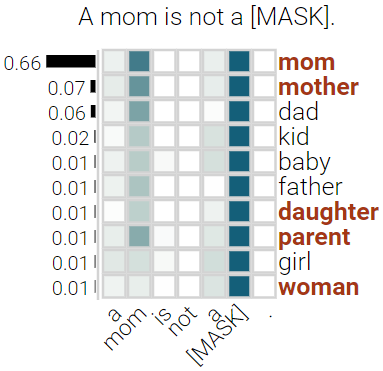}
\end{subfigure}%
\begin{subfigure}{.5\linewidth}
  \centering
  \includegraphics[scale=0.37]{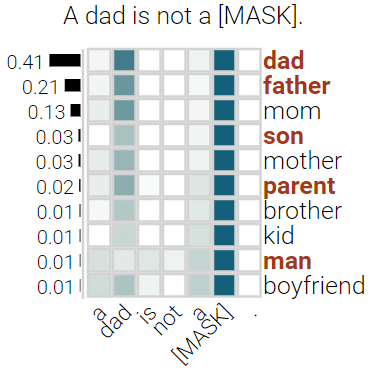}
\end{subfigure}
\caption{BERT: Similarity visualization of the first 10 predictions of the negation sentences \textit{A mom is not a [MASK] - A dad is not a [MASK].}}
\vspace{-1.5em}
\label{fig:syn_neg}
\end{figure}

\begin{figure}[!ht]
\centering
\begin{subfigure}{.5\linewidth}
  \centering
  \includegraphics[scale=0.4]{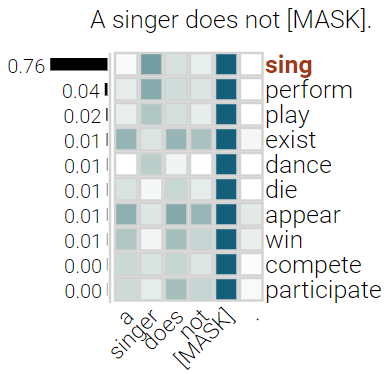}
\end{subfigure}%
\begin{subfigure}{.5\linewidth}
  \centering
  \includegraphics[scale=0.4]{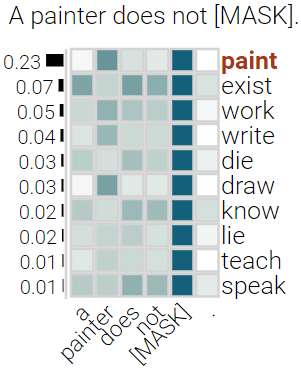}
\end{subfigure}
\caption{BERT: Similarity visualization of the first 10 predictions of the negation sentences \textit{A singer does not [MASK] - A painter does not [MASK].}}
\vspace{-0.5em}
\label{fig:syn_neg2}
\end{figure}

\paragraph{Forbidden predictions for semantic pairs:} 
\texttt{forbidden@1}~lies at 41\% for BERT,~19\% for ALBERT, and 32\% for RoBERTa.~\texttt{forbidden@5} is at 67\% for BERT, 53\% for ALBERT, and 54\% for RoBERTa, while \texttt{forbidden@10} lies at 73\% for BERT, 58\% for ALBERT, and 65\% for RoBERTa. For BERT and RoBERTa the coordination subset seems to be the hardest. The easiest for BERT is the syntactic negation, and for RoBERTa -- the lexical negation. In contrast, for ALBERT the easiest is the coordination subset, and the hardest is the lexical negation subset. These results might create the impression that BERT has the most difficulty in this task and that newer models such as ALBERT perform better. However, the further findings and the interpretation following shall shed light on this preliminary impression. Also, considering the embarrassingly easy examples included in our datasets, the amount of forbidden predictions is alarming for all datasets (see Figure~\ref{fig:syn_neg}, Figure~\ref{fig:syn_neg2}\footnote{Note that the sentence \textit{The painter does not paint} is not an impossible sentence, and thus an LM might have seen it. However, as noted in Section \ref{s:data}, LMs should have mainly learned prototypical notions, and thus such predictions should not occur within the 10 most probable ones.}).

\begin{figure}[!ht]
\centering
\begin{subfigure}{.5\linewidth}
  \centering
  \includegraphics[scale=0.4]{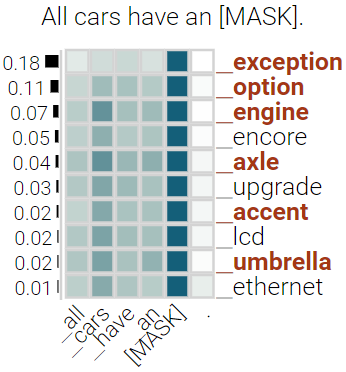}
\end{subfigure}%
\begin{subfigure}{.5\linewidth}
  \centering
  \includegraphics[scale=0.38]{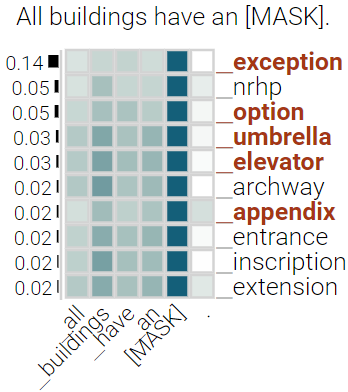}
\end{subfigure}
\vspace{-0.5em}
\caption{ALBERT: Similarity visualization of the first 10 predictions of the sentences \textit{All cars have an [MASK] - All buildings have an [MASK].}}
\vspace{-0.7em}
\label{fig:all_cars_build}
\end{figure}

\paragraph{[ALBERT] Similar predictions independently from context:} 
For negation and quantifiers, RoBERTa and ALBERT perform better than BERT. 
However, if we examine the examples more closely, we see that ALBERT predicts similar words no matter the exact sentence (see Figure \ref{fig:all_cars_build}). 
For the Inconsistent Dataset, all positive sentences have similar predictions to each other and all negative ones as well. 
It is doubtful that the training data contained such concordances and this raises the question if this model is contextualized in the same way BERT is.

\begin{figure}[!ht]
\centering
\begin{subfigure}{.5\linewidth}
  \centering
   \includegraphics[scale=0.4]{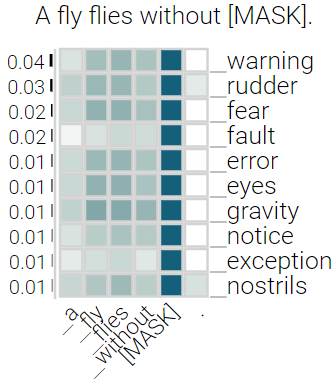}
\end{subfigure}%
\begin{subfigure}{.5\linewidth}
  \centering
   \includegraphics[scale=0.4]{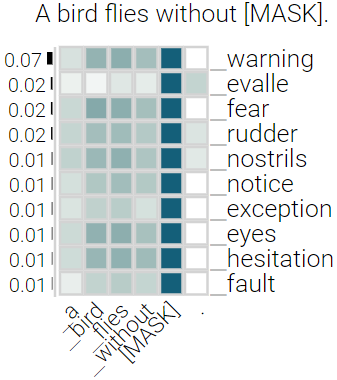}
\end{subfigure}
\vspace{-0.5em}
\caption{ALBERT: Similarity visualization of the first 10 predictions of the lexical negation sentences \textit{A fly flies without [MASK] - A bird flies without [MASK].}}
\vspace{-0.8em}
\label{fig:fly}
\end{figure}

\paragraph{[ALBERT, RoBERTa] Predictions correlate only with the predicate:} For the lexical negation subset of the Semantic Dataset, ALBERT's predictions are not based on the entire context, i.e., the predicate, the subject, and any function words, but rather correlate only with the predicate of the sentences. For example, all sentences containing the verb \textit{fly} lead to the same predictions, no matter whether it is a \textit{fly}, an \textit{owl} or a \textit{bird} flying (see Figure \ref{fig:fly}).  

\begin{figure}[!ht]
\centering
\begin{subfigure}{.5\linewidth}
  \centering
   \includegraphics[scale=0.4]{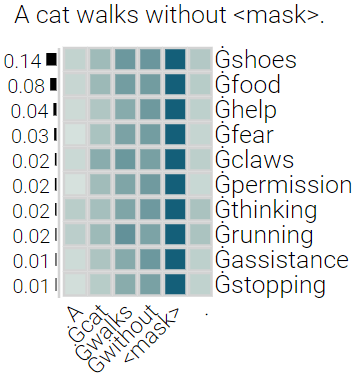}
\end{subfigure}%
\begin{subfigure}{.5\linewidth}
  \centering
   \includegraphics[scale=0.4]{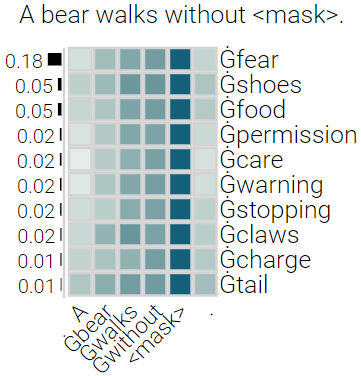}
\end{subfigure}
\vspace{-0.5em}
\caption{RoBERTa: Similarity visualization of the first 10 predictions of the lexical negation sentences \textit{A cat walks without <mask> - A bear walks without <mask>.}}
\vspace{-0.8em}
\label{fig:walks}
\end{figure}

Similar observations can be made for RoBERTa. 
It has been shown that LMs can learn stereotypical associations reasonably well, e.g., that \textit{walk} is related to \textit{shoes} as in the given example in \autoref{fig:walks}. 
These associations suggest that LMs are capable of learning commonsense reasoning -- knowledge accepted by the majority of people, e.g., how the world works \cite{Bhargava_Ng_2022}. 
Although the given examples suggest that the models are capable to learn such associations, the visualizations reveal that the predictions are stronger related to the sentence's predicate than the subject. The same goes for the predicates such as \textit{swim} or \textit{see}. 

\begin{figure} [!ht]
\centering
\begin{subfigure}{.5\linewidth}
  \centering
   \includegraphics[scale=0.33]{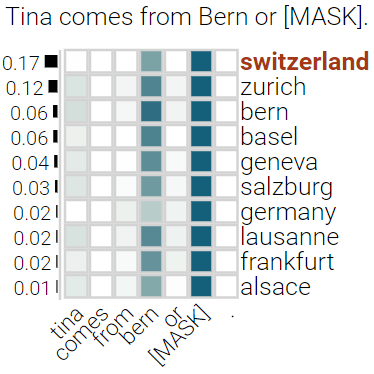}
\end{subfigure}%
\begin{subfigure}{.5\linewidth}
  \centering
   \includegraphics[scale=0.33]{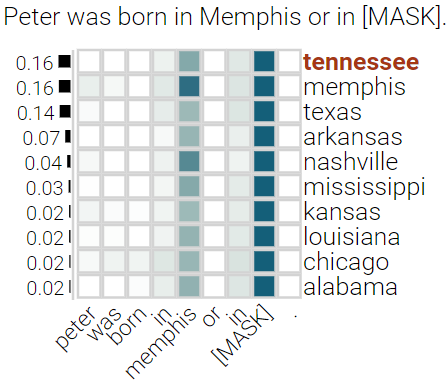}
\end{subfigure}
\caption{BERT: Similarity visualization of the first 10 predictions of the sentences \textit{Tina comes from Bern or [MASK] - Peter was born in Memphis or in [MASK].}}
\label{fig:loc}
\vspace{-1em}
\end{figure}

\paragraph{[BERT] Prediction quality dependent on specific named entities:} Taking a closer look at the coordination partition of the Semantic Dataset and the location examples specifically, we observe a high similarity between the named entity of the location already contained in the sentence and the predicted word, which is also a named entity in 100\% of the cases (see Figure \ref{fig:loc}). This not only indicates that the presence of named entities has a strong influence on the learning outcome because the remaining context is ignored, but also that the specific named entities have an impact on the predictions' quality. Particularly, some country/state-capital combinations lead to more forbidden predictions than others. For example, the US states \textit{Texas, California, Arizona, Florida} are more often within the \texttt{forbidden@1} than states such as \textit{Indiana, Tennessee, Minnesota}. This might suggest that the training data of BERT contained more instances of the former combinations, and the model learned a strong relation between these named entity pairs, 
ignoring any other (functional or lexical) words. 

\subsection{Insights and Interpretation}
In the following, we describe some potential reasons why current LMs fail in making linguistically correct word predictions.

\begin{figure}[!ht]
\centering
\begin{subfigure}{.5\linewidth}
  \centering
  \includegraphics[scale=0.4]{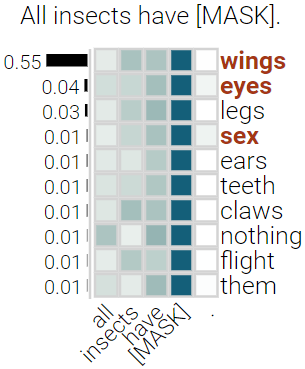}
\end{subfigure}%
\begin{subfigure}{.5\linewidth}
  \centering
  \includegraphics[scale=0.4]{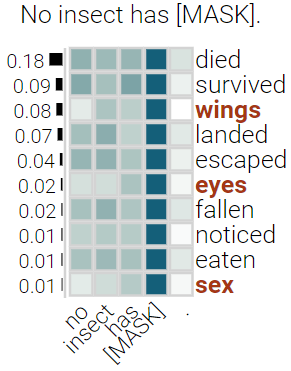}
\end{subfigure}
\caption{BERT: Similarity visualization of the first 10 predictions of the disjoint pair \textit{All insects have [MASK] - No insect has [MASK].}}
\label{fig:all_no}
\vspace{-0.5em}
\end{figure}

\paragraph{Function words ignored in semantically related contexts:}
We find that if the masked word has some semantic relation with the other main concept of the sentence -- most often the subject, then the functional word embeddings have low similarity to the masked word. In contrast, if no semantic relation can be established, there is a higher similarity to the function words. For example, in Figure \ref{fig:all_no}, we can see that all predictions that are related to (animals') body parts and have some semantic relation with the subject \textit{insects} do not exhibit any cosine similarity to the quantifiers, i.e., the first cell of the matrix is white, while for irrelevant predictions such as \textit{nothing, died, eaten} there is a similarity between the predicted word and the quantifier. This indicates that in many cases the predictions are dominated by semantically rich words and the model ignores other functional operators. 

\paragraph{Contextualization of function words:}
The visualizations show that if all words of the sentence have similar (high or low) similarities to the predicted word, then the predictions are mostly neither inconsistent nor forbidden -- even if they do not belong to the average common sense. This suggests that if the masked word is predicted considering the whole context, it should have a similarity to all words in the context. 
Related work supports this assumption \citep{ethayarajh-2019-contextual,sevastjanova-etal-2021-explaining}: functions words have high similarity to the rest of the words in higher layers because they are highly contextualized, i.e., they become very context specific. However, our findings show that this mode of learning leads to the negligence of function words: in some cases, the contextualization of function words is indeed high, i.e., their similarity to the predicted word is high, and then no inconsistent predictions arise. But when the sentence contains words that are semantically related, the model gets distracted, the function words are glossed over and their similarity to the predicted word drops. In consequence, their semantic nature is not captured. 

\begin{figure}[!ht]
\centering
  \includegraphics[width=\linewidth]{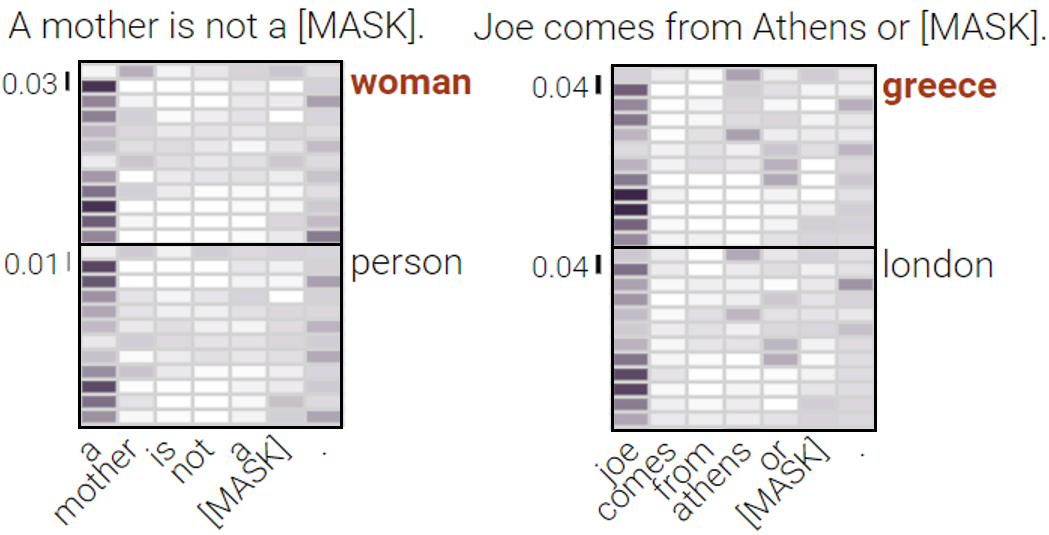}
\caption{BERT: Attention visualization of 2 predictions of the sentences \textit{A mother is not a [MASK]} and \textit{Joe comes from Athens or [MASK].} For each prediction, rows represent the layerwise average attention from the predicted word to the words (columns) in the sentence.}
\label{fig:attention}
\vspace{-0.5em}
\end{figure}

\paragraph{Attention confirms insights:}
Visualizing attention can show whether the observed behavior is specific to cosine similarity or can also be retraced in attention patterns. Indeed, we observe that throughout the layers the first word receives the most attention no matter its part-of-speech. Function words, such as \textit{not} and \textit{or} in Figure \ref{fig:attention}, receive higher attention only in the middle layers, which have been shown to mainly capture syntactic properties \citep{jawahar2019-bert-structure}. Thus, it seems that the semantic constraints of these words, which should be captured in the higher layers, are indeed left unaccounted for. 

\paragraph{Models are sensitive to minor input changes:}
While extracting predictions for RoBERTa model, we noticed its high sensitivity to the provided input. Even additional space(s) between tokens in the input context can produce different prediction probabilities, ranks, or even different predictions. Thus, we want to sensibilize researchers to be careful when working on similar experiments in order to avoid such potential, undesired errors.

\section{Conclusion}
This paper presented the extent to which LMs learn semantic constraints of function words. Based on theoretical linguistic literature, we created new datasets for testing three functional classes, and showed that popular masked LMs make problematic predictions. The visualizations in the developed web interface highlighted potential reasons for this poor performance. 
In this work, we set out to shed light on a subset of function word categories.
Future work shall look into other function word categories that ought to be challenging for LMs and uncover additional reasons for the poor model performances.

\section*{Acknowledgments}
We thank the Deutsche Forschungsgemeinschaft (DFG, German Research
Foundation) for funding within the project
BU 1806/10-2 ``Questions Visualized'' of the FOR2111 and project D02 ``Evaluation Metrics for Visual Analytics in Linguistics'' (Project ID: 251654672  -- TRR 161).

\bibliography{custom}

\newpage
\appendix
\section{Appendix A: Sample of Datasets}
\label{sec:appendixA}

\begin{table}[!ht]
\begin{small}
\begin{center}
\begin{tabular}{|c|c|c|}
  \hline
   &  & \textbf{Examples} \\ 
  \hline
  \multirow{15}{*}{\rotatebox[origin=c]{90}{\small{Inconsistent Dataset}}} & \multirow{10}{*}{\rotatebox[origin=c]{90}{\small{negation}}} & Cairo is not located in [MASK]. \\
 && Cairo is located in [MASK]. \\
 \cline{3-3}
&& A guitar does not have [MASK].	\\
&& A guitar has [MASK].  \\
\cline{3-3}
 && A chair has no [MASK]. \\
&& A chair has [MASK]. \\
 \cline{3-3}
&& Maria is a mother without a [MASK]. \\
&& Maria is a mother with a [MASK]. \\
 \cline{3-3}
&& A cat sees without [MASK].\\
&& A cat sees with [MASK]. \\
 \cline{2-3}
& \multirow{4}{*}{\rotatebox[origin=c]{90}{\small{coord.}}} & Joe is a dolphin or an [MASK].	\\
&& Joe is a dolphin and an [MASK].  \\
\cline{3-3}
&& Tina is a bird or an [MASK]. \\
&& Tina is a bird and an [MASK].\\
 \cline{2-3}
& \multirow{6}{*}{\rotatebox[origin=c]{90}{\small{quantifiers}}} & All cars have an [MASK].	\\
&& No car has an [MASK]. \\
 \cline{3-3}
&& All cooks [MASK].	\\
&& No cook [MASK]. \\
 \cline{3-3}
&& Some shoes have [MASK].\\
&& All shoes have [MASK].\\
 \cline{2-3}
  \hline
  \hline
\multirow{20}{*}{\rotatebox[origin=c]{90}{\small{Semantic Dataset}}} & \multirow{6}{*}{\rotatebox[origin=c]{90}{\small{syn. negation}}} & A mom is not a [MASK]. \\
&& \textbf{forbidden}: mom, mother, grandmother, \\ 
&& grandma, granddaughter, bride, wife, \\
&& woman, niece, stepmother, daughter, aunt	 \\
 \cline{3-3}
&& A designer does not [MASK]. \\
&& \textbf{forbidden}: design \\
 \cline{3-3}
&& A guitar does not have [MASK]. \\
&& \textbf{forbidden}: strings \\
\cline{2-3}
& \multirow{6}{*}{\rotatebox[origin=c]{90}{\small{lex. negation}}} & A bird flies without [MASK].\\
&& \textbf{forbidden}: wings \\
 \cline{3-3}
&& John is a father without a [MASK].\\
&& \textbf{forbidden}: child \\
\cline{3-3}
&& Peter is a brother without a [MASK].	\\
&& \textbf{forbidden}: sibling \\
\cline{2-3}
& \multirow{8}{*}{\rotatebox[origin=c]{90}{\small{coordination}}} & Mark was born in Athens or in [MASK]. \\
&& \textbf{forbidden}: Greece \\
\cline{3-3}
&& Tina died in Beijing or in [MASK].\\
&& \textbf{forbidden}: China \\
\cline{3-3}
&& George comes from Berlin or [MASK]. \\
&& \textbf{forbidden}: Germany \\
\cline{3-3}
&& Kate is a cat or an [MASK].\\
&& \textbf{forbidden}: animal \\
  \hline
\end{tabular}
\caption{Sample sentences from each dataset and each subset. The complete datasets can be found under: \url{https://function-words.lingvis.io/}.}
\label{tab:examples}
\end{center}
\end{small}
\end{table}

\end{document}